\documentclass[authorversion, sigconf, acmthm=false, nonacm=true]{acmart}

\AtBeginDocument{%
  \providecommand\BibTeX{{%
    \normalfont B\kern-0.5em{\scshape i\kern-0.25em b}\kern-0.8em\TeX}}}

\usepackage[capitalize]{cleveref}
\crefname{section}{Sec.}{Secs.}
\Crefname{section}{Section}{Sections}
\Crefname{table}{Table}{Tables}
\crefname{table}{Tab.}{Tabs.}

\usepackage{multirow}
\usepackage{float}
\usepackage{newfloat}
\usepackage{listings}
\usepackage{amsmath}
\usepackage{caption}
\usepackage{subcaption}
\graphicspath{ {images/} }
\floatstyle{ruled}
\newfloat{listing}{tb}{lst}{}
\floatname{listing}{Listing}

\usepackage[symbol]{footmisc}

\begin{document}

%%
%% The "title" command has an optional parameter,
%% allowing the author to define a "short title" to be used in page headers.
\title{SP-ViT: Learning 2D Spatial Priors for Vision Transformers}

%%
%% The "author" command and its associated commands are used to define
%% the authors and their affiliations.
%% Of note is the shared affiliation of the first two authors, and the
%% "authornote" and "authornotemark" commands
%% used to denote shared contribution to the research.
% \settopmatter{authorsperrow=1} %make the template consider one author per row
\newcommand{\tsc}[1]{\textsuperscript{#1}} %shorthand for superscripts
\author{Yuxuan Zhou\tsc{1}\footnotemark[1], Wangmeng Xiang\tsc{2}\footnotemark[1], Chao Li\tsc{4}, Biao Wang\tsc{4}, Xihan Wei\tsc{4}}
\author{Lei Zhang\tsc{2}, Margret Keuper\tsc{3}, Xiansheng Hua\tsc{4}} 
%to break line, start another author block
\author{\vskip .2cm}
\email{zhouyuxuanyx@gmail.com}
%   \vspace{1.5cm}
% \authornote{Work done as an intern at Alibaba Group.}
% \vspace{\linewidth}
% \affiliation{\vskip .2cm}
\affiliation{
\vspace{0.2cm}
  \institution{$^1$ University of Mannheim}
  \institution{$^2$ The Hong Kong Polytechnic University}
  \institution{$^3$ University of Siegen, Max Planck Institute for Informatics, Saarland Informatics Campus}
    \institution{$^4$ Alibaba Group}
}

\renewcommand{\shortauthors}{Zhou, et al.}

%%
%% The abstract is a short summary of the work to be presented in the
%% article.
\begin{abstract}
 Recently, transformers have shown great potential in image classification and established state-of-the-art results on the ImageNet benchmark. However, compared to CNNs, transformers converge slowly and are prone to overfitting in low-data regimes due to the lack of spatial inductive biases.
    Such spatial inductive biases can be especially beneficial since the 2D structure of an input image is not well preserved in transformers. 
    In this work, we present Spatial Prior–enhanced Self-Attention (SP-SA), a novel variant of vanilla Self-Attention (SA) tailored for vision transformers. 
    Spatial Priors (SPs) are our proposed family of inductive biases that highlight certain groups of spatial relations.  
    Unlike convolutional inductive biases, which are forced to focus exclusively on hard-coded local regions,
    our proposed SPs are learned by the model itself and take a variety of spatial relations into account. Specifically, 
    %% 拆成短句而不要用从句，很难看懂！！！！！！！两方面的东西不要放到同一个句子
    %%% 强调不同的inductive biases可以互补
    %%% 而不要强调 global ！！！！！！！！！！！！！！！！因为mlp什么的和sa都是global的
    %%%%改一下说法！！！！！！！！！！核心就是学习到了不同的sp！！！！！！
    %%% 全文包括ablation处都要围绕这个中心思想来！！！！！！ 
    the attention score is calculated with emphasis on certain kinds of spatial relations at each head,  
    and such learned spatial foci can be complementary to each other. 
    Based on SP-SA we propose the SP-ViT family, which consistently outperforms other ViT models with similar GFlops or parameters. Our largest model SP-ViT-L achieves a record-breaking 
    $86.3\%$ Top-1 accuracy with a reduction in the number of parameters by almost 50\% compared to previous state-of-the-art model (150M for SP-ViT-L$\uparrow$384 vs 271M for CaiT-M-36$\uparrow$384) among all ImageNet-1K models trained on $224\times224$ and fine-tuned on $384\times384$ resolution w/o extra data. 
\end{abstract}

\maketitle
 \pagestyle{plain}

\begin{figure}[t]
  \centering
  \includegraphics[width=1.1\linewidth]{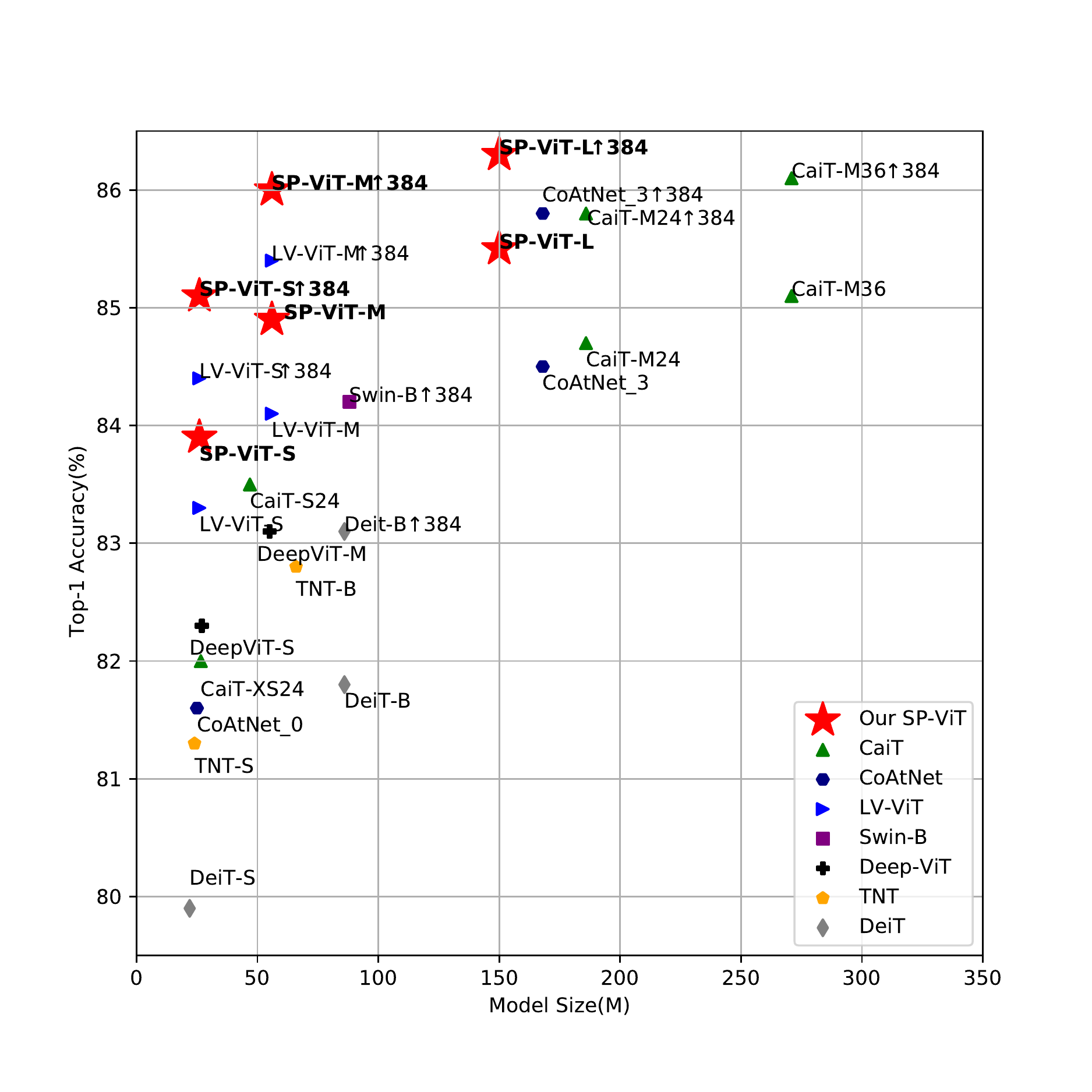} % Reduce the figure size so that it is slightly narrower than the column.
  \caption{ImageNet-1K top-1 accuracy of our proposed SP-ViT and state-of-the-art ViTs. 
  The models shown are all trained on $224\times224$ resolution, $\uparrow$ denotes that models are fine-tuned on a higher resolution.
  Note that we exclude models pretrained on extra data or larger resolution than $224\times224$ for a fair comparison.
%   Token Labeling \cite{jiang2021all} and Knowledge Distillation \cite{touvron2021training} are used as 
%   extra data augmentations for training our SP-ViTs and CaiTs respectively.
  } 
  \label{fig:intro}
\end{figure}

\begin{figure*}[t]
  \centering
  \begin{subfigure}[b]{0.56\textwidth}
      \includegraphics[scale=0.27]{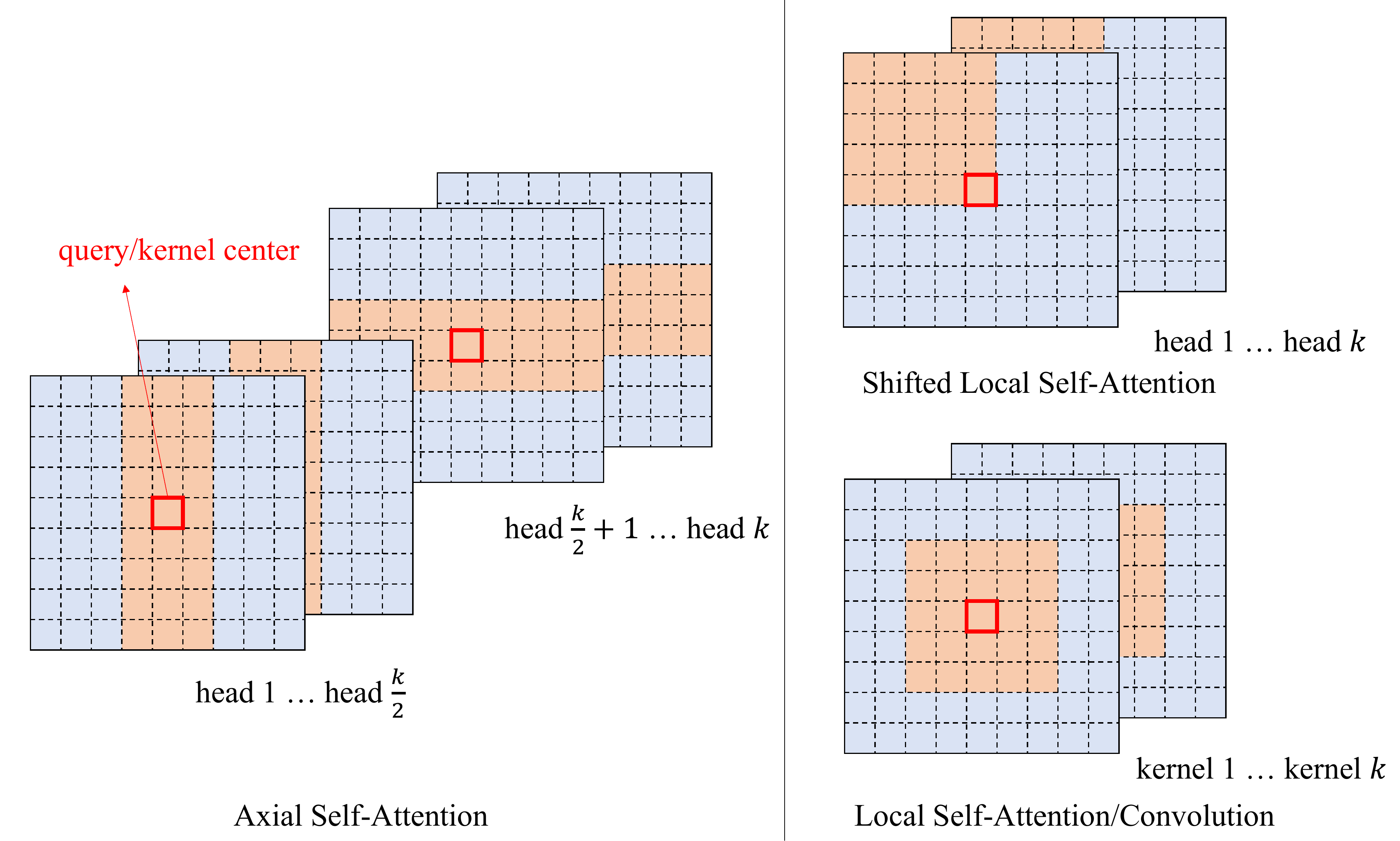}
      \caption{Hand-crafted convolutional inductive biases}
      \label{rfidtest_xaxis}
  \end{subfigure}
  \begin{subfigure}[b]{0.42\textwidth}
      \includegraphics[scale=0.27]{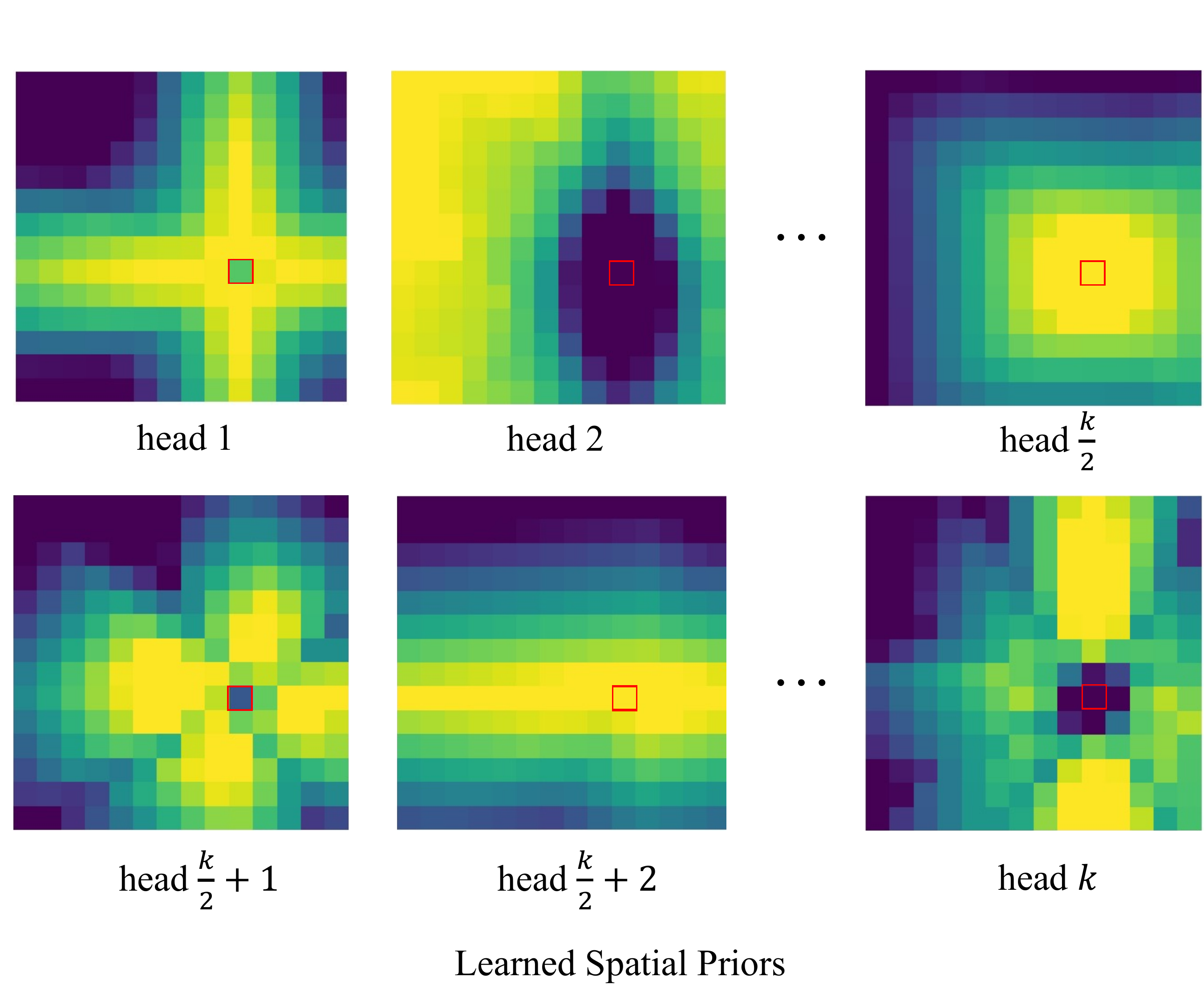}
      \caption{Proprosed SP-SA}
      \label{rfidtest_xaxis}
  \end{subfigure}
  \caption{(a) Examples of convolutional inductive biases proposed for ViTs: axial self-attention in CSWin-Transformer \cite{dong2021cswin} and shifted local self-attention in Swin-Transformer \cite{liu2021swin}.
  (b) The presented Spatial Priors (SPs) are learned by our model automatically. The learned SPs
  assign different scores for different spatial relations. Given a certain SP, attention is forced to be within high-score regions. 
  It can be seen that our SP-SA handles different types of spatial relations in a complementary manner, e.g., SPs which focus on local and non-local relations are both learned by our model. 
  Consequently, a more global receptive field can be maintained at a single layer.}
  % \cite{section ablation}
  \label{fig1}
\end{figure*}
\section{Introduction}
\label{sec:intro}

\footnotetext[1]{Work done as Intern at Alibaba Group}
Transformers \cite{vaswani2017attention} have recently achieved exciting results in image classification \cite{dosovitskiy2021an, liu2021swin, yuan2021tokens, han2021transformer, chen2021crossvit, touvron2021training, dong2021cswin, yuan2021incorporating, dai2021coatnet, wu2021cvt, touvron2021going, zhou2021deepvit, jiang2021all, d'ascoli2021convit}, after dominating in natural language processing (NLP) tasks \cite{devlin2018bert, liu2019roberta, brown2020language}. 
% competive results to convolutional neural networks(CNNs), cite some cnn papers?
At the heart of all transformers lies the so-called self-attention mechanism, which captures the content relations between all pairs of input tokens globally and focuses on related pairs selectively. 
% this is not related to our theme, but brings additional terms, unnecessary
% It indeed resembles one important feature of human cognition system, i.e., selective processing of incoming sensory information. 
% directly emphasize on the lack of prior, before this state that previous works hasn't considered prior information
% for example, we can first say that transformer lack certain types of inductive bias first
Self-attention is more flexible in comparison to convolution, which is hard-coded to capture local dependencies exclusively. This can possibly equip transformer models with larger capacity and greater potential for computer vision tasks. 
As reported in recent works, transformers outperform Convolutional Neural Networks (CNNs), when pretrained on large dataset \cite{dosovitskiy2021an}, facilitated with knowledge distillation \cite{touvron2021training} or pseudo labels \cite{jiang2021all} from pretrained CNNs.
% apart from selective attention, it is well established that prior information learned from past experiences facilitates the cognitive process as well.
% Although one might suffer from optical illusion under certain occasions, prior expectations or “inductive bias” are indispensable for a learner to extrapolates beyond the data, by constraining the space of models considered by the learner. \cite{tenenbaum2011how, michalski2013machine, lake2017building}

Nevertheless, CNNs generalize better and converge faster than Vision Transformers~(ViT). 
This suggests that certain types of inductive biases employed in convolution can still be beneficial to solving vision tasks.
Not surprisingly, many recent studies \cite{liu2021swin, dong2021cswin,yuan2021incorporating, dai2021coatnet, yuan2021tokens, wu2021cvt, graham2021levit,
touvron2021training, d'ascoli2021convit} propose to incorporate convolutional inductive biases into ViTs in different ways and all demonstrate
boosted performance. 
The effectiveness of convolution relies on the fact that neighboring pixels of natural images are highly correlated, 
but there may exist other highly correlated contents outside the local receptive field of a convolutional filter that are ignored. Therefore, we propose to 
make use of a variety of inductive biases simultaneously, just as humans do in our daily life, 
e.g., if we see a part of a horizontal object, we naturally look along its direction instead of restricting our sight within a local area.

% For these methods, the trade-off between local and global is inevitable, 

% It is indeed well established that inductive bias plays an important role in human cognitive process as well. 
% Although one might suffer from optical illusion under certain occasions, prior expectations or “inductive bias” are indispensable for a learner to extrapolates beyond the data, 
% by constraining the space of models considered by the learner \cite{tenenbaum2011how, michalski2013machine, lake2017building}.

In this work, we introduce a novel family of inductive biases named \emph{Spatial Priors} (SPs) into ViTs 
via an extension of vanilla self-attention (SA), called \emph{Spatial Prior–enhanced Self-Attention} (SP-SA). 
SP-SA highlights a certain group of 2D spatial relations at each attention head based on the relative position of the key and query patches. It facilitates the calculation of attention scores within the context of such types of spatial relations. 
Since the construction and validation of appropriate spatial priors are extremely laborious, we introduce the idea of \emph{learnable spatial priors}.
More specifically, we only impose the weak prior knowledge to the model that different relative distances should be treated differently.
Yet we do not force the model to favor any kind of spatial relation in advance, e.g., neither local nor non-local. Effective spatial 
priors (SPs) are supposed to be discovered by the model itself in the training stage. 
For this purpose, 
SPs are represented by a family of mathematical functions which map the relative coordinates to
abstracted scores, called \emph{spatial relation functions}. To search for desirable spatial relation functions, 
we parameterize these functions by neural networks and optimize them jointly with ViTs. 
Thereby, the model can learn spatial priors similar to the ones induced in convolutions, but it can also learn spatial relationships over larger distances. 

Examples for learned SPs are shown in \cref{fig1}(b). Diverse complementary patterns are presented in different attention heads, so that 
different types of spatial relations are handled individually. At the same time, a global receptive field is approximately composed by considering all heads together.

% The proposal of Spatial Priors is motivated by the observed gap between language and visual inputs: 
% the 2D spatial relations are no longer maintained 
% after an input image being divided into a sequence of patches, which are analogous to input tokens in Natural Language Processing(NLP).
% The content of an image is hardly comprehensible after flattening. 
% Because the inverse of flattening is unknown for the model, the problem remains unsolved despite that absolute positional embeddings are added to the input.
% Thus injecting some prior knowledge about spatial relations could probably bridge the gap and bring great advantages to ViTs. 

As a matter of fact, convolutional inductive biases can be seen as a special kind of spatial priors: 
they first divide coordinate spatial relations into two categories, i.e., ones focusing on the local neighborhood and ones focusing on non-local regions. 
Then they learn priors of the local neighborhoods and ignore the non-local relations. 
%imposes a strong prior on the model that the local region is to focus on and the non-local to ignore. 
For comparison, some of the existing approaches to combine such convolutional biases with ViTs are illustrated in \cref{fig1}(a).  
Based on a %the basic 
local window, new variants are proposed for CSWin-Transformer \cite{dong2021cswin} and Swin-Transformer \cite{liu2021swin} by changing the aspect ratio or shifting the center respectively.  
However, the design of such windows is extremely intuitive and the main idea of focusing on local relations remains unchanged.

In summary, we make the following contributions:
\begin{itemize}
\item We propose a family of inductive biases for ViTs 
that focus on different types of spatial relations, called Spatial Priors (SP). SPs generalize the
locally restricted convolutional inductive biases
to both local and non-local correlations. Parameterized with neural networks, SPs are automatically learned
during training, without imposing a preference for any hard-coded region in advance.

\begin{figure*}[t]
  \centering
  \includegraphics[scale=0.5]{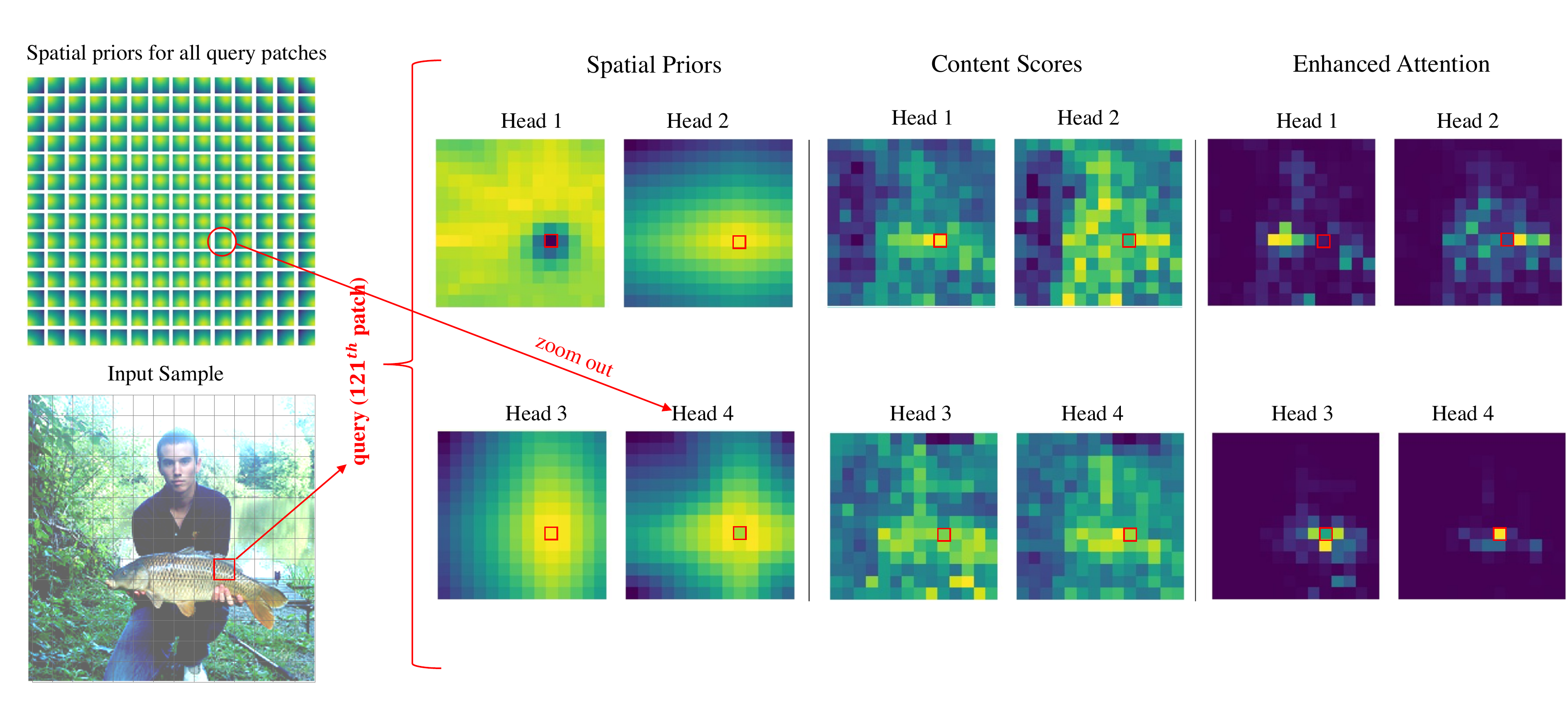} % Reduce the figure size so that it is slightly narrower than the column.
  \caption{Visualization of the learned 2D SPs, content scores and the enhanced attention at the first 4 heads of layer 4. The input image is shown
  in the bottom-left and the query patch is marked in red. Different SPs are learned, including
  horizontal and vertical (head 2 and 3), non-local (head 1), as well as cross-shaped (head 4).
  The attention scores at each head are obtained within the context of a certain type of spatial relations. The original attention is distracted by background objects, whereas our Spatial Priors help the model to focus on the object of interest.} 
  \label{fig3}
\end{figure*}

\item We propose SP-SA, a novel self-attention variant that automatically learns 
beneficial 2D spatial inductive biases. Built on our proposed SP-SA, we further construct a ViT variant called SP-ViT.
SP-ViTs achieve state-of-the-art results on the ImageNet Benchmark w/o extra data, e.g., SP-ViT-L$\uparrow$384 yields a record-breaking $86.3\%$
Top-1 accuracy with only around half the amount of parameters of the previous best-performing model CaiT-M-36$\uparrow$384. 

\item Our proposed SPs 
% are beneficial for general vision tasks. SPs are 
are compatible with various input sizes, as they are derived from relative coordinates between each pair of patches instead of their absolute positions. SP-ViTs also demonstrate improved performance over the baseline model on Image Classification when fine-tuned on higher resolution. 
% and on downstreaming tasks such as segmentation.

% \item We alleviate the dilemma of local convolutional inductive bias vs global vanilla self-attention. 
% Complementary patterns highlighting different regions are presented in the learned SPs at different heads,
% so a more global receptive field can be maintained. 
%We hope this insight may draw more attention of the research community to finding appropriate inductive biases for vision transformers.

%SP-ViTs achieve the state-of-the art performance on ImageNet1k classification task.
%% 参考cait
\end{itemize}

% \begin{figure*}[ht]
%   \centering
%   \includegraphics[scale=0.6]{Model Architecture.pdf} % Reduce the figure size so that it is slightly narrower than the column.
%   \caption{The schema of SP-ViT. SP-SA can be used as a drop-in replacement for the vanilla SA layer at a range of depths.
%   Because the classification token does not have a valid 2D relative coordinate, it is simply concatenated with the hidden representation after the last SP-SA layer.
%   FFN: feedforward network (2 linear layers separated by a GeLU activation).
%   }% \cite{section ablation}
  
%   \label{fig2}
%   \end{figure*}

\section{Related Work}
\label{sec:related}

\subsection{Vision Transformers}

Recently, Dosovitskiy et al.~\cite{dosovitskiy2021an} showed that purely attention-based transformers can achieve state-of-the-art performance in image classification, when pre-trained on large-scale datasets. 
Since then, a vast amount of efforts have been made to improve ViTs. 
Some works \cite{jiang2021all, gong2021improve} find it effective to add additional losses or regularization terms, 
while others propose new patch embedding blocks \cite{han2021transformer} or scale-up methods \cite{touvron2021going, zhou2021deepvit}. 
\cite{liu2021swin, dong2021cswin, zhang2021aggregating, wang2021pyramid} propose to utilize multi-scale information, where pyramid designs and local attention are often adopted to reduce the overall computation.   
It is noteworthy that an approximately cross-shaped 2D structure, which brings performance gain in CSWin-Transformer \cite{dong2021cswin}, is also learned automatically by our model. 
%From other perspectives, a  are also suggested.

\subsection{Inductive Biases for Vision Transformers}
The performance of ViTs degrades rapidly with a reduced amount of training data.
To alleviate this issue, many studies focus on emphasizing local correlations by introducing a convolutional inductive bias into ViTs,
either by restricting self-attention to local windows \cite{liu2021swin, dong2021cswin}, 
combining vanilla transformers with implicit or explicit convolutional operations 
\cite{yuan2021incorporating, dai2021coatnet, yuan2021tokens, wu2021cvt, graham2021levit}, knowledge distillation~\cite{touvron2021training}, or convolutional initialization \cite{d'ascoli2021convit}.
Our work also incorporates inductive biases into ViTs, but they are not locally restricted and automatically learned by the model. 
Indeed, as shown in \cref{fig1}(b), patterns that focus solely on local or remote regions are both present in the learned~SPs. 

\subsection{Relative Spatial Information for ViTs}
Transformers are by their very nature permutation invariant, thus extra spatial information is often supplied to make them more suitable for ordered input data. 
Besides the commonly adopted absolute positional embedding, the relative spatial information
is also considered in Swin-Transformer \cite{liu2021swin} by an additional trainable
bias term called relative positional bias. ConViT \cite{d'ascoli2021convit} also
introduces a function taking coordinates relative to the query patch as input to force the attention to be within a local region.  
In comparison to ViTs, using relative positional information is more common in NLP transformers.
The relative positional embedding \cite{shaw2018self} is built on the 1D distances between tokens and has been further improved in XL-Net \cite{yang2019xlnet} and DEBERTA \cite{he2021deberta}.
However, the original relative positional embedding can be extended to 2D for ViTs with little effort. 
In contrast, our method not only provides 2D spatial information but also imposes learned priors about spatial relations on the model. The latter proves to be beneficial for learning a stronger model, see \cref{sec:ablation}.

\begin{figure*}[ht]
  \centering
  \includegraphics[scale=0.65]{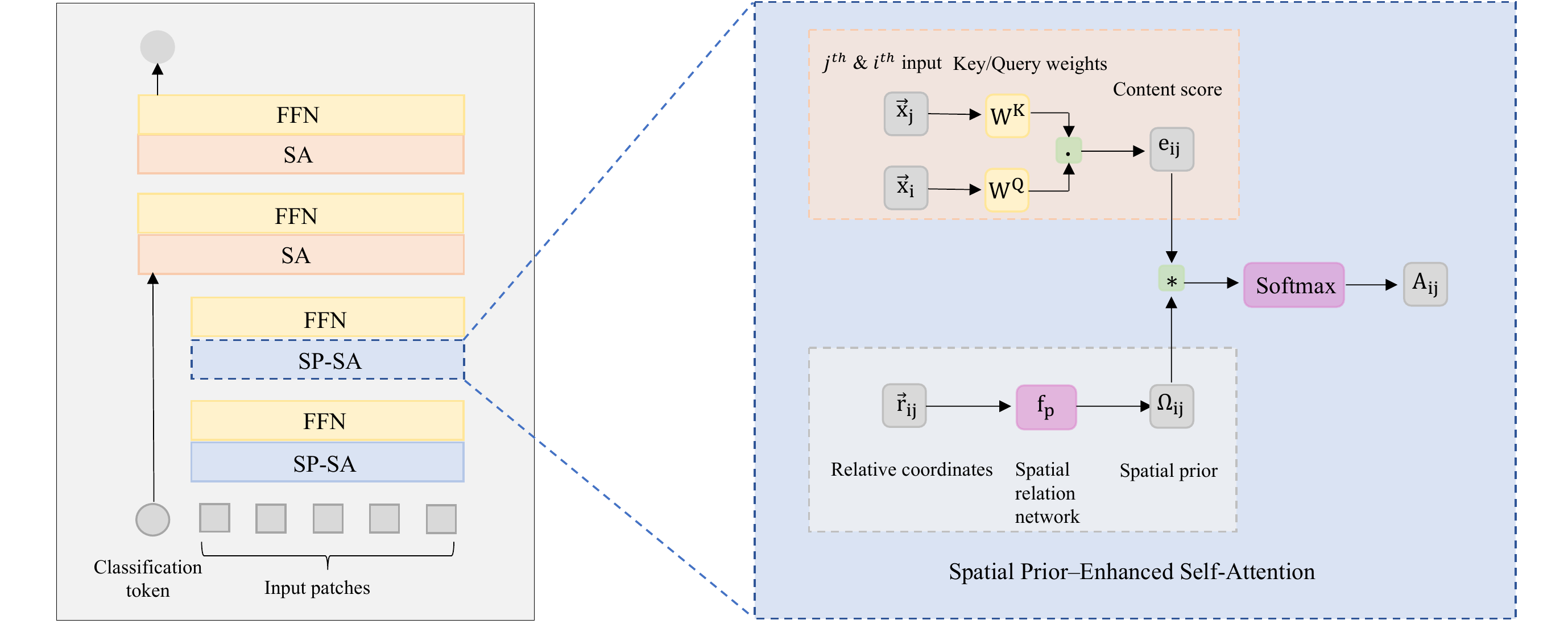} % Reduce the figure size so that it is slightly narrower than the column.
  \caption{The schema of SP-ViT. SP-SA can be used as a drop-in replacement for the vanilla SA layer at a range of depths.
  Because the classification token does not have a valid 2D relative coordinate, it is simply concatenated with the hidden representation after the last SP-SA layer.
  FFN: feedforward network (2 linear layers separated by a GeLU activation).
  }% \cite{section ablation}
  
  \label{fig2}
  \end{figure*}
  
% \begin{figure*}[t]
%   \centering
%   \includegraphics[scale=0.5]{Visualization.pdf} % Reduce the figure size so that it is slightly narrower than the column.
%   \caption{Visualization of the learned 2D SPs, content scores and the enhanced attention at the first 4 heads of layer 4. The input image is shown
%   in the bottom-left and the query patch is marked in red. Different SPs are learned, including
%   horizontal and vertical (head 2 and 3), non-local (head 1), as well as cross-shaped (head 4).
%   The attention score at each head is obtained within the context of a certain type of spatial relations. The original attention are distracted by background objects, whereas our Spatial Priors help the model to focus on the object of interest.} 
%   \label{fig3}
% \end{figure*}

\section{Method}
\label{sec:method}

\subsection{Revisiting Multi-Head Self-Attention}

% describe the functionality of attention: assigning weight to elements
Self-attention receives an input sequence $\mathbf{X}=(\vec{x}_1,...,\vec{x}_n)$ and outputs a new sequence $\mathbf{Y}=(\vec{y}_1,...,\vec{y}_n)$ of the same length, where each element is computed as weighted sum of the linearly transformed input elements:
\begin{equation}
\vec{y}_i = \sum_{j=1}^n A_{ij}(\vec{x}_j^{\top}\mathbf{W}^V)
,
\end{equation}
each weight coefficient, or attention score $A_{ij}$, is determined based on the semantic dependencies between two elements, by applying a softmax function to
the scaled dot product of linearly transformed elements $e_{ij}$, which we refer to as content score throughout this paper:

% {\rm exp}
\begin{align}\label{att} 
 A_{ij} &=\frac{{\rm exp} (e_{ij})} {\sum^n_{k=1} {\rm exp} (e_{ik})},
 \end{align}
 with
 \begin{align}
%\end{equation}
%
%\begin{equation}
e_{ij} &= \frac{(\vec{x}_i^\top\mathbf{W}^Q) (\vec{x}_j^\top\mathbf{W}^K)^\top}{\sqrt{d_z}}.
\end{align}

Multi-head self-attention employs several such operations in parallel to learn different kinds of interdependencies. 
The final output is obtained by applying a linear transformation to the concatenated outputs from each head.

%% query patch 字体太粗了，改细一点
%%% 把它改成一个head一个head， 然后 Spatial Prior x content score softmax attention score的形式！！！！！！

% Self-attention extracts semantic dependencies between the elements of an input sequence to obtain the so-called attention scores. Then the  
% aggregates the information into the output sequence 

% then describe how, query key, value mappings, cos similarity
% then equation

\subsection{Spatial Prior-enhanced Self-Attention}

Motivated by the observation that certain types of inductive biases on spatial relations can be beneficial to transformers,
we propose an extension of self-attention enhanced by a combination of learned 2D Spatial Priors (SPs), called Spatial Prior–enhanced Self-Attention (SP-SA).

Each SP $\mathbf{\Omega} \in \mathbf{R}^{N \times N}$ forms a kind of spatial context for computing attention scores $\mathbf{A}\in \mathbf{R}^{N \times N}$, 
and it is derived from coordinate spatial relations between input element pairs,
i.e.~relative positions between the key and query patches for ViTs.
Thus an SP has exactly the same form of attention scores and we can simply integrate it in \cref{att} by multiplicative interaction:

\begin{equation}
    % a_{ij} = {\rm softmax}(e_{ij}*\omega_{ij})
    A_{ij} = \frac{{\rm exp} (e_{ij} \cdot \Omega_{ij})} {\sum^n_{k=1} {\rm exp} (e_{ik} \cdot \Omega_{ik})}.
\label{sp-sa}
\end{equation}

\subsubsection{2D Spatial Prior}
Taking query patch $i$ as the reference point, we can obtain a relative coordinate $\vec{r}_{ij} \in \mathbf{R}^{2}$ for image patch $j$. 
Then we employ a shared mapping $f_p$ for all query and key patch pairs, named spatial relation function:

\begin{equation}
    \Omega_{ij} = f_p(\vec{r}_{ij}),
\end{equation}
the outputs together form the so-called 2D SP Matrix $\Omega$.

\subsubsection{Learnable 2D Spatial Priors}
To enable the model to learn desirable inductive biases automatically, we 
employ Multilayer Perceptron (MLP) to parameterize the mapping from 2D relative coordinates to $\mathbf{\Omega}$:
\begin{equation}
    f_p(\vec{r}_{ij}) = \mathrm{MLP}(\vec{r}_{ij}).
\end{equation}

Thereby, we allow $\Omega$ to learn a weighting for the attention scores for query $\vec{x}_i$ and key $\vec{x}_j$ which depends solely on their relative coordinates and is applied in a non-linear way, i.e.~before the softmax.  
We extend SP-SA to its multi-head version by adding a unique network to each head. 
%It shares the same spirit of muti-head self-attention 
This design follows the same motivation as 
%This is motivated in a similar way as 
multi-head self-attention and assumes
that a combination of different SPs should boost the performance. 

\subsection{Relation to Other Methods}
In the following, we discuss the similarities and differences of SP-SA to the most closely related works in detail.
\subsubsection{Relation to Local Windows}
The square and cross-shaped windows used in \cite{liu2021swin, dong2021cswin} can be seen as a special form of our proposed spatial relation functions in practice:
% equation is used for one-line equation
% use amsmath split

\begin{align}
           f_p(\vec{r}_{ij}) = 
    \begin{cases}
            1, \quad & \text{if } \| (\vec{r}_{ij} - \vec{\Delta}) \odot (a, b)  \|^{\infty} <= 1 \\
            \quad &\text{or } \| (\vec{r}_{ij} - \vec{\Delta}) \odot (b, a)\|^{\infty} <= 1 , \\
            0, \quad & \text{else }
    \end{cases}
\label{window}
\end{align}

\noindent where $\vec{\Delta}$, $a$ and $b$ control the shift, window width and height respectively. If $a=b$, it generates a square window, otherwise it results in a cross-shaped window.
Both works only adopt some hard-coded patterns for the whole network, while our method proposes to benefit from a variety of 
beneficial 2D structures.

\subsubsection{Relation to PSA}

The Positional Self-Attention (PSA) proposed by  d’Ascoli et al. \cite{d'ascoli2021convit} can also be regarded as a manually designed family of spatial relation functions:
\begin{equation}
    f_p(\vec{r}_{ij}) = \alpha (\|(\Delta_x, \Delta_y)\|^2-\| \vec{r}_{ij} - (\Delta_x, \Delta_y)\|^2),
\label{local.init}
\end{equation}

\noindent where $\Delta_x \text{ and } \Delta_y$ are learnable parameters, 
which are initialized following an explicit rule to approximate convolution effect.

Note that their main contribution is the so-called local/convolutional initialization, 
which restricts the number of heads to the square of integer numbers, and the initial values of both $\alpha$ and $\vec{\Delta}$ require extra hyperparameter tuning. 
In order to compare with their method, we adopt a ViT baseline with 9 heads for our ablation analysis as in \cite{d'ascoli2021convit}.
Moreover, it may be possible to extend their proposed local initialization to our SP-SA, 
leveraging the merits of both methods. We leave this to future work.

\subsubsection{Relation to Relative Positional Embeddings}
Shaw et al. \cite{shaw2018self} introduce the so-called 1D relative positional embedding for transformers to take relative distances into account:
\begin{equation}
    % a_{ij} = {\rm softmax}(e_{ij}*\omega_{ij})
    A_{ij} = \frac{{\rm exp}(e_{ij}+(\vec{x}_i\mathbf{W}^Q)T_{r_{ij}})} {\sum^n_{k=1} {\rm exp} (e_{ik}+(\vec{x}_i\mathbf{W}^Q)T_{r_{ij}})},
\label{rpe}
\end{equation}

\noindent where $T$ is a learnable embedding table from which the relative positional embedding is taken. Then it interacts multiplicatively with the query.

Unlike 1D relative positional embeddings, our proposed spatial relation functions have more power in modeling the 2D structures of images. 
If extended to 2D, their method is equivalent to applying a linear transformation to one-hot representations of relative distances. 
For one-hot representations, the magnitude of distances is neglected, while this is not the case for relative coordinates.  
More importantly, Shaw et al. \cite{shaw2018self} simply add relative positional embeddings to key and value in the attention equation. The modification of self-attention 
is rather intuitive but lacks a clear physical interpretation. As opposed to positional embeddings,
our method not only provides neutral positional information but also learns and injects beneficial inductive biases into the model.

%To show the advantages of our approach, we have discussed the relations and differences to some closely related works in detail.
The proposed SP-SA offers more flexibility than the above discussed approaches. We also confirmed experimentally %found out 
that our proposed SA variant outperforms these methods, see \cref{table.5c}.

 \begin{table}[h]
         \caption{Comparing to state-of-the-art models trained on ImageNet-1k $224\times224$ resolution.  
        Models are by default trained and tested on $224\times224$ resolution if not specified. 
        $\uparrow$ plus size denotes the model is trained on $224\times224$ resolution then fine-tuned and tested on $\text{size}\times\text{size}$ resolution. The performance of LV-ViT-L trained on $224\times224$ resolution is not available in \cite{jiang2021all}. And  LV-ViT-L trained on $288\times288$ resolution has a lower accuracy of $85.3\%$.}
        \centering
        %\resizebox{.95\columnwidth}{!}{
        \begin{tabular}{lccc}
           \toprule
            Network     & Top-1 (\%) &  Parameters & FLOPs \\
            \midrule
           DeiT-S \cite{touvron2021training} & 79.9 &  22M& 4.6B\\
     
            DeepViT-S \cite{zhou2021deepvit} & 82.3 & 27M & 6.2B\\
      
           T2T-ViT-14 \cite{yuan2021tokens} & 81.5 & 22M & 5.2B \\
      
           TNT-S \cite{han2021transformer} & 81.3 & 24M & 5.2B \\
           CaiT-XS-24 \cite{touvron2021going} & 82.0 & 27M & 5.4B \\
            LV-ViT-S \cite{jiang2021all} &83.3& 26M & 6.6B \\ 
            EfficientNet-B5 \cite{tan2019efficientnet} & 83.6 & 30M & 9.9B\\
             \textbf{Our SP-ViT-S} &\textbf{83.9}&  26M& 6.6B \\ 
           \midrule
         
            DeiT-B  \cite{touvron2021training} & 81.8  & 86M& 17.5B   \\
            T2T-ViT-24 \cite{yuan2021tokens} & 82.3 & 64M & 14.1B \\
        ConViT-B \cite{d'ascoli2021convit}& 82.4 & 86M & 17.5B \\
         TNT-B \cite{yuan2021tokens} & 82.8 & 66M & 14.1B \\
          
         Swin-S \cite{liu2021swin} & 83.0 & 50M & 8.7B \\
        
           DeepViT-L \cite{zhou2021deepvit} & 83.1 & 55M &12.5B \\
            
           Swin-B \cite{liu2021swin}& 83.3 & 88M& 15.4B \\
      CaiT-S-24 \cite{touvron2021going} & 83.5 & 47M & 9.4B \\
    %   CSWin-S \cite{dong2021cswin} & 83.6 & 35M & 6.9B \\
    ConvNeXt-B \cite{liu2022convnet} & 83.8 & 89M &  15.4B \\
        EfficientNet-B5 \cite{tan2019efficientnet} & 84.0 & 43M & 19.0B\\
           LV-ViT-M \cite{jiang2021all}& 84.1 & 56M & 12.7B \\
        %    LV-ViT-L$\uparrow \uparrow$288& \multicolumn{1}{c}{85.7} & \multicolumn{1}{c}{150M} & 59.0B \\

            %   CSWin-B \cite{dong2021cswin} & 84.2 & 78M & 15.0B \\
                 \textbf{Our SP-ViT-M} & \textbf{84.9} & 56M  & 12.7B \\
              \midrule
                    ConvNeXt-L \cite{liu2022convnet} & 84.3 & 198M & 34.4B \\
            CoAtNet-3 \cite{dai2021coatnet} & 84.5 & 168M & 34.7B \\ 
           CaiT-M-24 \cite{touvron2021going}& 84.7 & 186M & 36.0B \\
    
           \textbf{Our SP-ViT-L} &  \textbf{85.5}& 150M  & 34.7B  \\ 
          
        \midrule
        T2T-ViT-14$\uparrow$384 \cite{yuan2021tokens} & 83.3 & 22M & 17.1B  \\
        CoAtNet-0$\uparrow$384 \cite{dai2021coatnet} & 83.9 & 25M & 13.4B \\
       LV-ViT-S$\uparrow$384 \cite{jiang2021all} & 84.4 & 26M & 22.2B \\
            \textbf{SP-ViT-S$\uparrow$384}  & \textbf{85.1} & 26M & 22.2B \\
        \midrule 
      Swin-B$\uparrow$384  \cite{liu2021swin}& 84.2 & 88M& 47.0B \\

           CaiT-S-24$\uparrow$384 \cite{touvron2021going}&85.1 &47M & 32.2B \\
           LV-ViT-M$\uparrow$384 \cite{jiang2021all} & 85.4& 56M & 42.2B \\
                % CSWin-B$\uparrow$384  \cite{dong2021cswin} & 85.4 & 78M & 47.0B \\
                \textbf{Our SP-ViT-M$\uparrow$384} &  \textbf{86.0}& 56M & 42.2B \\
            
           \midrule
           CoAtNet-3$\uparrow$384 \cite{dai2021coatnet} & 85.8 & 168M & 107.4B \\
            CaiT-M-24$\uparrow$384 \cite{touvron2021going}&85.8 &186M & 116.1B \\
            CaiT-M-36$\uparrow$384 \cite{touvron2021going} & 86.1 & 271M & 173.3B \\ 
           \textbf{Our SP-ViT-L$\uparrow$384}  &  \textbf{86.3} & 150M  & 110.6B \\ 
           \bottomrule
        \end{tabular}
        % 把cite的文献放到说明这里
        \label{table4}
        \end{table}
        
\section{Experiments}
We first provide an experimental evaluation of the proposed SP-SA in the context of image classification on the ImageNet-1k dataset and show that SP-ViTs achieve state-of-the-art results for training without extra data. Further, we provide an extensive ablation study to analyze the impact of all proposed model details.

\begin{figure*}[h]
  \centering
  % i used 12 fontsize in ppt
  \includegraphics[scale=0.7]{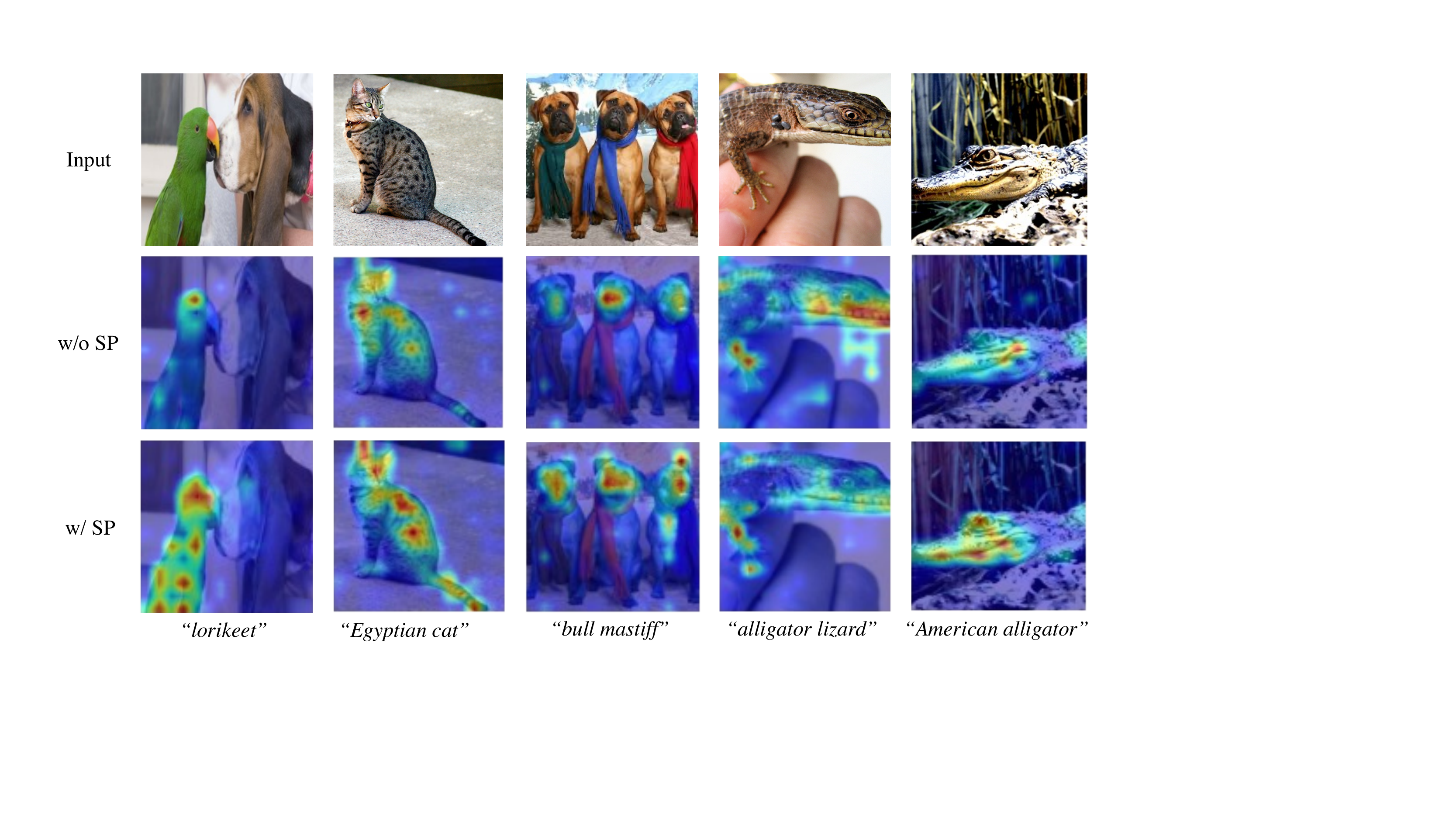} % Reduce the figure size so that it is slightly narrower than the column.

  \caption{Visualization using Transformer Explainability~\cite{Chefer_2021_CVPR}. The second row are results of DeiT baseline, which has no SP layers. The Last row are results of SP-ViT. Our SP-ViT generate results which have more focus on areas of interests and less distraction from background comparing to DeiT.}
  % \cite{section ablation}
  \label{vis_spvit}
  \end{figure*}

\subsection{Image Classification on ImageNet-1K}
   
\subsubsection{Settings}

        % \begin{table*}[ht]
        %     \centering
        %     %\resizebox{.95\columnwidth}{!}{
        %     \begin{tabular}{l|cccccc|c}
        %            \hline
        %         \multirow{2}*{Method} & \multicolumn{6}{c|}{Additional Spatial Relation} &\multirow{2}*{Top-1 acc (\%)}\\
           
        %             % \cline{2-8}
        %         &  \multicolumn{1}{l}{Multiplicative}  &  \multicolumn{1}{l}{Spatial Focus} & \multicolumn{1}{l}{(Quasi-)Global} & \multicolumn{1}{l}{Multi-Head} & \multicolumn{1}{l}{Nonlinear} &\multicolumn{1}{l|}{Ordered} &   \\
        %        \hline
        
        %    SP Only &- & $\checkmark$& $\checkmark$& $\checkmark$&$\checkmark$&$\checkmark$&\\
        %     MLP-Mixer                                                                                                 \\
          
        %         \hline
        %     \end{tabular}
        %     \label{table6}
        %     \caption{Comparing 
        %     %     ).
        %     }
        % \label{t.function}
        % \end{table*}
        
All models for ImageNet-1K classification are trained on a single machine node with 8 Tesla V100 GPUs.
Our code is based on PyTorch \cite{paszke2019pytorch} and the official implementation of DeiT \cite{touvron2021training}.
To obtain our SP-ViT model, 
we replace the vanilla SA layers of the DeiT baseline with our proposed SP-SA till the last 2 layers and follow the default training settings in \cite{jiang2021all} (data augmentations,
training schedule, hyperparameters such as batch size, total number of layers etc.). 
We keep the vanilla SA in the last 2 layers, based on the preliminary ablation analysis conducted on a fraction of ImageNet. 
Please refer to \cref{sec:ablation} for more details.
When fine-tuning on $384\times384$ resolution (indicated by $\uparrow$384 in \cref{table4}), we set the batch size to 512, the learning rate to 5e-6, the
weight decay to 1e-8 and we fine-tune the model for 30 epochs. For more details please refer to the Appendix.

\subsubsection{Comparing to State-of-the-Art Models}
We compare our proposed SP-ViT (taking LV-ViT as our baseline) with other recent ViTs in \cref{table4},
including DeiT \cite{touvron2021training}, TNT \cite{han2021transformer}, T2T \cite{yuan2021tokens}, Swin Transformer \cite{liu2021swin}, DeepViT \cite{zhou2021deepvit},
LV-ViT \cite{jiang2021all}, Swin Transformer \cite{liu2021swin}, CaiT \cite{touvron2021going}, etc.. The models in \cref{table4} are grouped according to comparable model sizes in terms of their number of parameters. Within all groups, the proposed SP-ViT outperforms competing models in terms of Top-1 accuracy. Our best result with Top-1 accuracy of 86.3\% is achieved with the SP-ViT-L$\uparrow$384 model. It outperforms all previous models while only having about 150M parameters as compared to 271M parameters of the second best model CaiT-M-36$\uparrow$384. Also note that our smaller model SP-ViT-M$\uparrow$384 already achieves a Top-1 accuracy of 86.0\% which is on par with CaiT-M-36$\uparrow$384 while reducing the amount of parameters from 271M to 56M (i.e. by a factor of about 4.8). 

In the following, we first analyze our results in terms of class activation visualizations and then show ablation studies on our model components.

\subsubsection{Qualitative results}

We present visualizations of target class activation maps using the recent Transformer Explainability~\cite{Chefer_2021_CVPR} for several images  in Figure~\ref{vis_spvit} to showcase the behavior of SP-ViT. While the DeiT model only shows class activations on small parts of the target class regions, for example on the head of the ``Lorikeet", the fur of the ``Egyptian cat" or the jaw of the ``American alligator", the proposed SP-ViT model shows class activations on wider target class regions. Thereby, it follows well class specific image regions such as the pointy ears as well as the tail of the ``Egyptian cat", and the dogs' ears in the ``Bull mastiff" class. The ``Alligator lizard" example as well as the ``American alligator" further show a significantly diminished class activation in background regions compared to DeiT. In summary, we make two observations: 1) The results that are generated by SP-ViT focus more on areas of target class objects comparing to DeiT. In ``Lorikeet", ``Bull mastiff", ``Egyptian cat" and ``American alligator", SP-ViT's activation maps clearly have a better coverage of target class; 2) The distraction by background is better suppressed, e.g.~in ``Alligator lizard", which leads to a cleaner activation map.

\subsection{Ablation Analysis}
\label{sec:ablation}
\subsubsection{Settings}
For the ablation study,
we employ a small DeiT model as the  baseline with 12 layers, 9 heads and 432 total number of embedded dimensions.
The choice of the number of heads is simply for a fair comparison with other related methods, because Positional Self-Attention (PSA) introduced by d’Ascoli et al. \cite{d'ascoli2021convit} requires such specific numbers (square of integer numbers) of heads.
Due to limited available computation resources, we 
train all model variants on the first 100 classes of ImageNet-1K called ImageNet-100 for 300 epochs, following the setup in \cite{d'ascoli2021convit}.
In this section, we simply take the accuracy at the last epoch for all models. 
This should be a fair comparison, since we adopt the same hyperparameters for different models without tuning.  
For all experiments in this section, we train the models on 4 NVIDIA P100 GPUs and adopt a batch size of 256. 
The rest of settings are kept the same as DeiT's w/o knowledge distillation in \cite{touvron2021training}.

\subsubsection{Numbers of Substituted SA Layers}

\begin{figure}[h]
  \centering
  \begin{subfigure}[12 layer SP-ViT]{0.23\textwidth}
      \includegraphics[scale=0.52]{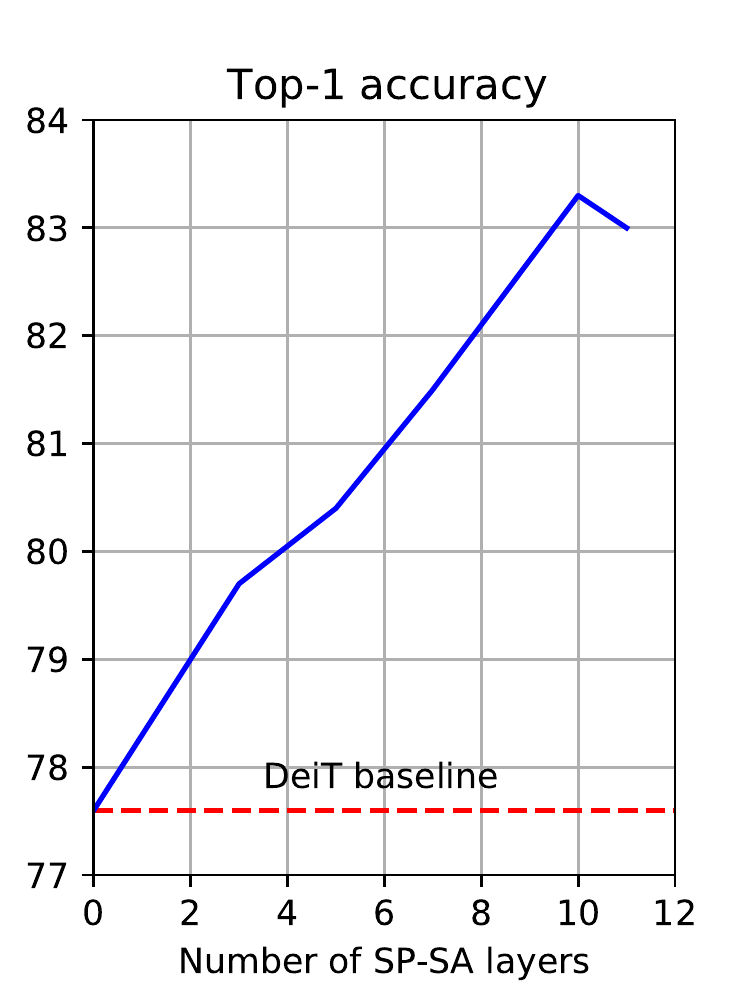}
      \caption{12 layer SP-ViT}
      \label{12_layer}
  \end{subfigure}
  \begin{subfigure}[16 layer SP-ViT]{0.23\textwidth}
      \includegraphics[scale=0.52]{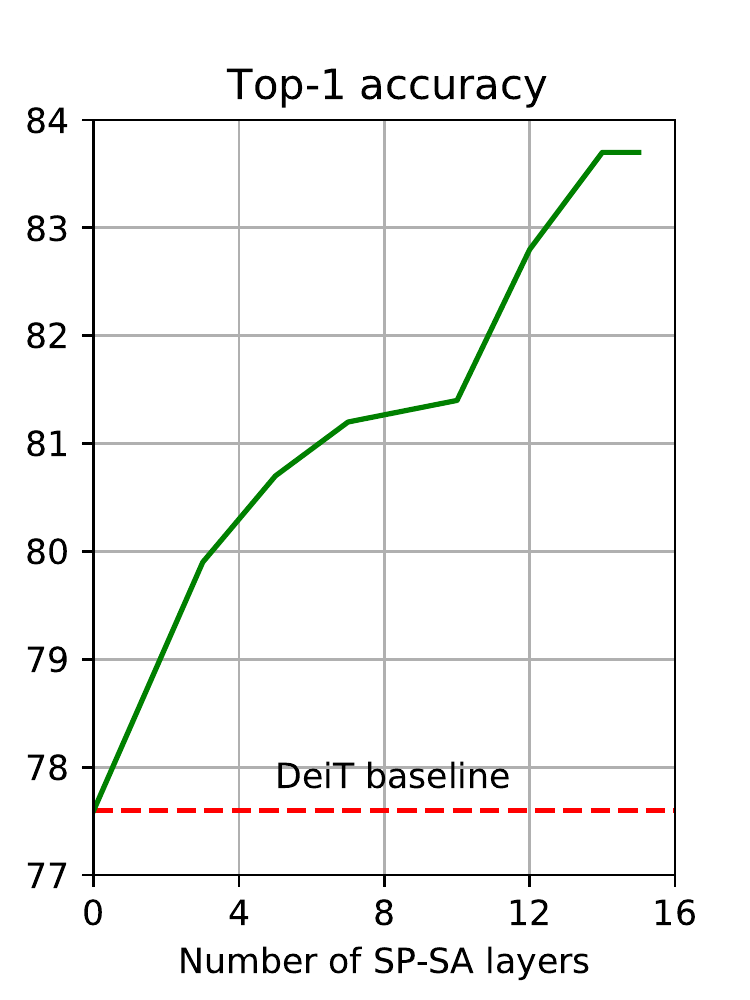}
      \caption{16 layer SP-ViT}
      \label{16_layer}
  \end{subfigure}

  \caption{Top-1 accuracy(\%) of SP-ViT on ImageNet-100 with different numbers of SP-SA layers. \cref{12_layer} and \cref{16_layer} show consistent improvements of our SP-ViT over DeiT Baselines with a total number of 12 and 16 layers respectively.}
  % \cite{section ablation}
%   \label{figlayer}
  \end{figure}
  
    % Note that when all 12 SA layers are replaced by SP-SA, the classification token no longer exists, so in this case we use Global Average Pooling (GAP) instead.
  
% \begin{table}[ht]
%   \caption{Eliminate the effect of inserting the classification token at later layers on ImageNet-100.}
%   \centering
%   %\resizebox{.95\columnwidth}{!}{
%   \begin{tabular}{ccc}
%      \hline
%      Sub. layers & Top-1 (\%) & Top-1 (\%)  \\
%      \hline
%       7   & 81.5  &  81.2 \\
%       10  & 83.3  &  81.0 \\
%       14  &   &  83.7 \\
%       15  &   &  83.7 \\

%      \hline
     
%   \end{tabular}
%   \label{table.compare}
%   \end{table} 

\begin{table}[ht]
  \caption{Eliminate the effect of inserting the classification token at later layers on ImageNet-100.}
  \centering
  %\resizebox{.95\columnwidth}{!}{
  \begin{tabular}{ccc}
     \toprule
     Sub. layers  & Cls token insertion layers & Top-1 (\%)  \\
     \midrule
     0 & 0  &   77.6 \\

     0 & 10  &   81.7 \\

     10 & 10&    \textbf{83.3} \\
     \midrule

     0 & Global Average Pooling&    79.5 \\

     12 & Global Average Pooling&    \textbf{81.7} \\

     \bottomrule
  \end{tabular}
  % 把cite的文献放到说明这里
  \label{table.compare}
  \end{table}

We first investigate how the performance of the SP-ViT is affected by the number of SP-SA layers. The layers are substituted from bottom to top of the model and a classification token is inserted after the last SP-SA layer.

It is shown in \cref{12_layer} that substituting a number of SA layers with SP-SA results in improved Top-1 accuracy comparing to DeiT baseline (0 layer). In general, the performance improves as more layers are substituted. For a model with 12 layers in total, the best performance is achieved when 10 layers are substituted. When substituting all but the last SA layer with our SP-SA, the performance drops slightly.
We hypothesize that this is very likely because the classification token is only involved in the last layer, thus the class-specific features are not adequately extracted in such a case. We further investigated models with more layers in \cref{16_layer} (16 layers in total), and found the similar trend. The best performance is achieved when the first to the penultimate layer are substituted. 

% When substituting 11 SA layers with SP-SA, the classification token is only involved in the last layer. This is very likely the reason for the slightly worse performance than substituting 10 layers.
% When all 12 SA layers are replaced by SP-SA, there does not exist any classification token, thus we use Global Average Pooling instead.  

As we have discussed in the section \ref{sec:method}, 
we add the classification token directly after SP-SA layers 
because there is no valid 2D relative coordinate for it. 
In addition, to exclude the influence of inserting the classification token at deeper layers instead of the first, 
we conduct a further comparison in \cref{table.compare}. 
It can be seen that the late insertion of the classification token does have a positive effect on
the classification result, which is also observed by Touvron et al. \cite{touvron2021going}. 
However, as aforementioned, inserting the classification token after SP-SA layers is a natural design choice for our SP-ViT 
and the substitution of SP-ViT also increases the accuracy further.

%%% 这个标题不要了，拆成3个小标题！！！！！！！！！！
%%% table 3那个打勾就不要了
%% table4 的打勾保留并改进

\begin{table}[ht]
  \caption{The effect of different hidden dimensions for our 2-layer MLP spatial relation function. Best results can be achieved with hidden dimension 32.
  }
  \centering
  %\resizebox{.95\columnwidth}{!}{
  \begin{tabular}{c|c}
         \toprule  
          % \cline{2-8}
          Hidden Dimension &   Top-1 acc (\%) \\
     \midrule
 16 & 83.2 \\ 
 32 & \textbf{83.6} \\ 
  64 & 83.3 \\
   128 & 83.1 \\ 
     %hai xuyao gen bucong weizhizuobiao xuechulai, ershi zhijie zuowei canshu xuechulai de qubie \\
     \bottomrule
  \end{tabular}
\label{table.mlp}
\end{table}
\subsubsection{Hidden Dimension of the 2-Layer MLP for the Spatial Relation Function}

\begin{table}[ht]
  \caption{Comparing to Self-Attention with Relative 
  Positional Bias \cite{liu2021swin}, Positional Self-Attention\cite{d'ascoli2021convit}, Self-Attention with 1D Relative 
  Positional Embedding (RPE) \cite{shaw2018self} and with its 2D extension as well as the 2D extension of a more advanced version of RPE proposed in DEBERTA \cite{he2021deberta} on ImageNet-100.
  }
  \centering
  %\resizebox{.95\columnwidth}{!}{
  \begin{tabular}{l|c}
         \toprule     
          % \cline{2-8}
          Method &   Top-1 acc (\%) \\
     \midrule

     1D RPE \cite{shaw2018self} &80.1 \\ 

     2D RPE \cite{shaw2018self}  & 79.9\\ 
     
     Improved 2D RPE \cite{he2021deberta}  & 82.8\\ 
     Relative Positional Bias \cite{liu2021swin} & 81.3 \\ 
     Positional Self-Attention \cite{d'ascoli2021convit} & 82.5 \\ 
 SP-SA Additive  & 83.5 \\ 
 SP-SA   & \textbf{83.6} \\ 
     %hai xuyao gen bucong weizhizuobiao xuechulai, ershi zhijie zuowei canshu xuechulai de qubie \\
     \bottomrule
  \end{tabular}
\label{table.5c}
\end{table}

\begin{table*}[ht]
  \caption{Comparing to extended methods of introducing relative spatial information on ImageNet-100:
  Local SA (7x7 window), Relative Positional Coefficient (RPC) (analogous to Relative Positional Bias \cite{liu2021swin}), 
  and SP-SA Linear (using a linear spatial relation function).
  }
  \centering
  %\resizebox{.95\columnwidth}{!}{
  \begin{tabular}{l|ccccc|c}
         \toprule
      \multirow{2}*{Method} & \multicolumn{5}{c|}{Interaction with Relative Spatial Information} &\multirow{2}*{Top-1 acc (\%)}\\
 
          % \cline{2-8}
      & \multicolumn{1}{l}{Spatial Focus} & \multicolumn{1}{l}{(Quasi-)Global} &  \multicolumn{1}{l}{Unique per Head} & \multicolumn{1}{l}{Nonlinear} &\multicolumn{1}{l|}{Ordered} &   \\
     \midrule

     Local SA &$\checkmark$&- &-&-&-& 81.0 \\
   
     RPC &$\checkmark$ &$\checkmark$& $\checkmark$ &-&-&  81.9  \\ % need to rerun c initialized as 1 rerun
     SP-SA Linear   & $\checkmark$ &$\checkmark$&$\checkmark$&-&$\checkmark$& 82.2 \\ 

SP-SA  & $\checkmark$& $\checkmark$& $\checkmark$&$\checkmark$&$\checkmark$& \textbf{83.6} \\

     %hai xuyao gen bucong weizhizuobiao xuechulai, ershi zhijie zuowei canshu xuechulai de qubie \\
     \bottomrule
  \end{tabular}
\label{table.9}
\end{table*}

We found that a simple 2-layer MLP works well as our spatial relation function. In \cref{table.mlp}, we investigate the impact of its hidden dimension on the performance. As we can see, the performance improves from a hidden dimension of 16 to 32 by $0.4\%$ in Top-1 accuracy, but a further increase in the hidden dimension does not further improve results. Results for a hidden dimension of 32 are the best.

\subsubsection{Comparing to Relative Positional Embedding}

%%% 那就把表格拆成n个小表格，然后这些打勾的项就不要了放到文章里讲，简要讲它们怎么做的，好在哪里为什么
SP-SA Additive is obtained by simply replacing the multiplication in \cref{sp-sa} with a summation. 
%% 这里直接讲出数字得了，更清楚，加和乘两项对比不用专门一个表格了
This setting is inferior to the original SP-SA, see \cref{table.5c}, but it is more comparable to other existing methods which also 
employ additive interaction between spatial information and content scores. 
To validate the effectiveness of our method, we compare it to several existing SA variants that also consider relative spatial information. 
As shown in \cref{table.5c}, our SP-SA has much higher Top-1 accuracy, compared to SA with 1D RPE and 2D RPE. 
Most importantly, relative Positional Embeddings do not impose any beneficial inductive bias to the vanilla SA as our SP. Further, its linear form is not capable of 
capturing sophisticated 2D spatial relations.
In addition, the value of $\vec{x}W^Q$ in \cref{rpe} changes dynamically for every different input query.
If an appropriate interaction between the content query and the positional embedding is always supposed to be achieved, a sufficiently large capacity is indispensable. 
However, this is not the case for the simple 
multiplication of query and relative positional embedding.

\subsubsection{Comparing to Relative Positional Bias}
As opposed to our method, the Relative Positional Bias \cite{liu2021swin} directly adds a univariate bias term to the content score between query-key pairs before applying softmax, 
and the bias term is taken from a learnable parameter table based on the relative coordinates.
Adding such a bias term is a straightforward idea to include relative spatial information, 
but it is neither based on the idea of nor capable of learning complex abstracted 2D spatial priors,  
as reflected in the results in \cref{table.5c}.     

\subsubsection{Comparing to Positional Self-Attention}
To compare our proposed learnable spatial relation function with the hand-crafted Positional Self-Attention \cite{d'ascoli2021convit} in \cref{local.init},
we have also compared SP-SA to Positional Self-Attention in \cref{table.5c}. Our method delivers better performance in this case of training a small model on limited data.
This shows that the effort in such a manual design process can be saved by our learnable SP.

\subsubsection{Importance of the Spatial Relation Function}
To further validate the importance of our proposed spatial relation function, we compare it to different ways of multiplicative interaction
with relative spatial information in \cref{table.9}. SP-SA Linear is obtained by removing the Relu of the default 2-layer Multi-Layer-Perceptron (MLP), 
and RPC is constructed by replacing our spatial relation function directly with coefficients taken from a parameter table, analogous to Relative Positional Bias \cite{liu2021swin}. 
For Local SA, we calculate attention scores only within a 7x7 local window. 
This is simply approximated by adopting \cref{window} with $a=b=1/3$ and $\vec{\Delta}=(0,0)$.
The results in \cref{table.9} indicate that powerful non-linear functions are necessary to learn
effective spatial priors.

%%% 暂时不用加进去！！！！！直接在ImageNet1k跑了跟resmlp比
% Moreover, the satisfying accuracy when using Spatial Priors alone supports their power as well. 

\subsubsection{Single vs Multiple Spatial Priors}

\begin{table}[h]
  \caption{The effect of unique Spatial Priors (SPs) per head. This setting performs better than unique SPs per layer as well as shared SPs.
  }
  \centering
  %\resizebox{.95\columnwidth}{!}{
  \begin{tabular}{lc}
     \toprule
     SP-SA &  Top-1 (\%)    \\
     \midrule
    shared SP            &   82.6 \\
     unique SPs per layer &   82.1 \\
     
unique SPs per layer\&head (default) &  \textbf{83.6} \\
     %hai xuyao gen bucong weizhizuobiao xuechulai, ershi zhijie zuowei canshu xuechulai de qubie \\
     \bottomrule
  \end{tabular}
  % 把cite的文献放到说明这里
  \label{table7}
  \end{table}

To validate the benefit of combining various SPs,
we compare our SP-SA to two of its variants: 
one only adopting a single SP for each layer, the other learning the same SP for the whole network. 
The results are reported in \cref{table7}. A shared SP for the whole network thereby provides better results than learning a single SP for each layer. However, the proposed setting with a unique learnable SP per layer$\&$head performs best, providing evidence of the benefit of combining different SPs.

\section{Conclusions and Discussions}

In this paper, we introduce a variant of Vanilla self-attention (SA) named Spatial Prior-enhanced Self-Attention (SP-SA) to facilitate vision transformers with automatically learned spatial priors.
Based on the SP-SA, we further proposed SP-ViT and experimentally demonstrate the effectiveness of our method. Our proposed SP-ViTs of different sizes all establish state-of-the-art results for models trained on ImageNet-1K only. For example, SP-ViT-M achieves a $0.8\%$ higher accuracy comparing to the previous state-of-the-art LV-ViT-M.  
We hope that our powerful SP-SA can stimulate more studies on designing and utilizing appropriate inductive biases
for vision transformers. Finally, the learned spatial priors can be binarized and used for designing more computation-efficient
transformers.

\noindent\textbf{Limitations} In spite of the excellent performance, vision transformers rely on heavy data augmentation techniques and the attention mechanism is computationally inefficient in comparison to convolution. Further investigations on improving ViTs are still desired.

\clearpage

%%
%% The next two lines define the bibliography style to be used, and
%% the bibliography file.
\bibliographystyle{ACM-Reference-Format}
\bibliography{sample-base}

%%
%% If your work has an appendix, this is the place to put it.
\appendix

% \section{Research Methods}

% \subsection{Part One}

% Lorem ipsum dolor sit amet, consectetur adipiscing elit. Morbi
% malesuada, quam in pulvinar varius, metus nunc fermentum urna, id
% sollicitudin purus odio sit amet enim. Aliquam ullamcorper eu ipsum
% vel mollis. Curabitur quis dictum nisl. Phasellus vel semper risus, et
% lacinia dolor. Integer ultricies commodo sem nec semper.

% \subsection{Part Two}

% Etiam commodo feugiat nisl pulvinar pellentesque. Etiam auctor sodales
% ligula, non varius nibh pulvinar semper. Suspendisse nec lectus non
% ipsum convallis congue hendrerit vitae sapien. Donec at laoreet
% eros. Vivamus non purus placerat, scelerisque diam eu, cursus
% ante. Etiam aliquam tortor auctor efficitur mattis.

% \section{Online Resources}

% Nam id fermentum dui. Suspendisse sagittis tortor a nulla mollis, in
% pulvinar ex pretium. Sed interdum orci quis metus euismod, et sagittis
% enim maximus. Vestibulum gravida massa ut felis suscipit
% congue. Quisque mattis elit a risus ultrices commodo venenatis eget
% dui. Etiam sagittis eleifend elementum.

% Nam interdum magna at lectus dignissim, ac dignissim lorem
% rhoncus. Maecenas eu arcu ac neque placerat aliquam. Nunc pulvinar
% massa et mattis lacinia.

\end{document}